\documentclass[twoside]{article}

\usepackage[letterpaper,inner=0.95in,outer=0.95in,top=0.85in,bottom=0.9in,includeheadfoot]{geometry}
\usepackage[numbers,sort&compress]{natbib}
\usepackage{palatino,epsfig,latexsym}
\usepackage{amsmath,amssymb}
\usepackage{algorithm}
\usepackage{algpseudocodex}
\usepackage{array}
\usepackage{graphicx}
\usepackage{adjustbox}
\usepackage{subcaption}
\usepackage{multirow}
\usepackage{float}
\usepackage{rotating}
\usepackage{placeins}
\usepackage{hyperref}

\newcommand{\TableFont}{\footnotesize\setlength{\tabcolsep}{2pt}\renewcommand{\arraystretch}{1.0}}

\begin{document}

\title{Robust Differential Evolution via Nonlinear Population Size Reduction and Adaptive Restart: The ARRDE Algorithm}

\author{
\parbox{0.94\textwidth}{\centering\small
Khoirul Faiq Muzakka$^{1,*}$, Ahsani Hafizhu Shali$^{2,3}$, Haris Suhendar$^{4}$,\\
S\"oren M\"oller$^{1}$, Martin Finsterbusch$^{1}$\\[0.5ex]
$^{1}$Institute of Energy Materials and Devices (IMD-2), Forschungszentrum J\"ulich GmbH, Germany\\
$^{2}$Research Center for Nuclear Physics, The University of Osaka, Japan\\
$^{3}$National Research and Innovation Agency, Indonesia\\
$^{4}$Department of Physics, Universitas Negeri Jakarta, Indonesia\\
$^{*}$Corresponding author: \texttt{k.muzakka@fz-juelich.de}
}
}

\maketitle
\begin{abstract}
		Robustness across heterogeneous optimization regimes remains a central challenge in bound-constrained continuous optimization. In practice, users often prefer optimizers that remain reliable across different dimensionalities, landscape structures, and evaluation budgets. Yet many Differential Evolution (DE) variants that perform strongly in one regime degrade substantially when transferred to others. To address this issue, we propose the \textit{Adaptive Restart--Refine Differential Evolution} (ARRDE) algorithm, a DE variant designed explicitly for cross-regime robustness. ARRDE combines an adaptive restart--refine strategy, a nonlinear population-size reduction schedule that depends on problem dimensionality, and a budget-aware population-initialization rule for restricted-budget settings.
	
	Because robustness cannot be established credibly from a narrow experimental setting, we evaluate ARRDE on five benchmark suites: CEC2011, CEC2017, CEC2019, CEC2020, and CEC2022. These suites span markedly different dimensions, landscape characteristics, and evaluation budgets, making this, to the best of our knowledge, one of the most comprehensive robustness-oriented evaluations reported for a proposed DE variant in this context. Since their official performance metrics emphasize different aspects and are not directly comparable, we additionally introduce a bounded accuracy-based scoring metric derived from relative error for cross-suite robustness assessment. Using both the official suite-specific metrics and the proposed unified metric, ARRDE demonstrates consistently strong performance and one of the most stable aggregate profiles across the five suites. These results support ARRDE as a competitive DE variant for robust optimization across heterogeneous benchmark regimes.
\end{abstract}



\section{Introduction}

Differential Evolution (DE)~\citep{Storn1997} is one of the most successful evolutionary algorithms for continuous optimization and has consistently produced top-performing variants in the IEEE Congress on Evolutionary Computation (CEC) competitions~\citep{BILAL2020103479}. Its effectiveness stems from a simple but powerful population-based search mechanism, together with a design that does not require gradient information and is therefore well suited to non-differentiable, non-convex, and noisy objective functions. Over the past two decades, many DE variants have been proposed, often by introducing adaptive control of the mutation factor $F$, crossover rate $CR$, and population size in order to improve convergence and solution quality.

For practical users, however, high performance in a single benchmark regime is not enough. In many real optimization tasks, the problem characteristics, dimensionality, landscape structure, and available evaluation budget are not known in advance to favor a particular algorithmic design, and extensive retuning is often infeasible. In such settings, an optimizer that performs reliably across substantially different regimes is generally more valuable than one that is highly specialized for a single benchmark family.

This robustness remains a major challenge in DE research. Recent studies have shown that many high-performing DE variants degrade substantially when evaluated outside the benchmark suites or evaluation budgets for which they were originally tuned~\citep{PIOTROWSKI2023101378,PIOTROWSKI2025101807}. In other words, methods that perform exceptionally well in one regime may perform poorly in another. This behavior is not entirely surprising: many advanced algorithms are developed, calibrated, or implicitly shaped under relatively narrow benchmark conditions, so some degree of specialization is expected. The cross-regime difficulty is particularly visible in the CEC benchmark literature. For example, the CEC2017 suite includes higher-dimensional problems ($D=10,30,50,100$) under the moderate budget $N_{\max}=10^4D$, whereas CEC2020 focuses on lower-dimensional problems ($D=5,10,15,20$) under much larger budgets of $5\times10^4$, $10^6$, $3\times10^6$, and $10^7$ evaluations, respectively. Algorithms that perform exceptionally well on CEC2017 often perform poorly on CEC2020, and vice versa. A similar pattern has also been reported across varying evaluation budgets, where methods that excel under small budgets may deteriorate sharply when the budget is large, while large-budget methods struggle when evaluations are limited~\citep{PIOTROWSKI2025101807}. A straightforward response would be to tune algorithms across many benchmark settings at once, but that may simply produce a compromise method that trades gains in one regime for losses in another. Our goal in this work is different. Rather than constructing a zero-sum compromise, we aim to design a DE variant whose search mechanisms remain genuinely competitive across heterogeneous optimization regimes. These observations motivate the need for DE variants that are robust not only to landscape differences, but also to systematic changes in dimensionality and computational budget.

To address this robustness problem, we propose the \textit{Adaptive Restart--Refine Differential Evolution} (ARRDE) algorithm, a DE variant designed explicitly to remain effective across contrasting optimization regimes. ARRDE combines an adaptive restart--refine mechanism, a nonlinear dimension-dependent population reduction schedule, and a budget-aware population-initialization rule. Because the central claim is robustness across regimes rather than performance in a single benchmark setting, the method is evaluated on five benchmark suites together with additional varying-budget experiments.

The main contributions of this work are as follows:
\begin{itemize}
	\item We propose ARRDE, a DE variant that aims to remain effective across contrasting optimization regimes, including high-dimensional moderate-budget, low-dimensional large-budget, and restricted-budget settings.
	\item We introduce an adaptive restart--refine mechanism, including final refinement and local exclusion during restart, to recover exploration after contraction while preserving intensification around promising regions.
	\item We introduce a nonlinear, dimension-dependent population reduction schedule that changes the exploration--exploitation balance across problem settings and works jointly with restart--refine to improve cross-regime robustness.
	\item We introduce a budget-aware population-initialization rule that scales the starting population with the available normalized evaluation budget, improving practicality in restricted-budget settings while remaining compatible with standard large-budget regimes.
	\item We conduct a broad robustness-oriented empirical study, including five CEC benchmark suites and additional experiments over varying normalized budgets $N_{\max}/D$, and we introduce a unified robustness-oriented metric for cross-suite comparison when official competition metrics are not directly comparable.
\end{itemize}

The study covers 212 problem instances: 22 application-motivated problems from CEC2011~\citep{cec2011}, 29 functions from CEC2017~\citep{cec2017} tested at four dimensions (10D, 30D, 50D, 100D), 10 functions from CEC2019~\citep{cec2019}, 10 functions from CEC2020~\citep{cec2020} tested at four dimensions (5D, 10D, 15D, 20D), and 12 functions from CEC2022~\citep{cec2022} tested at two dimensions (10D and 20D). In addition, ARRDE is tested on CEC2017, CEC2020, and CEC2022 under a wide range of normalized evaluation budgets $N_{\max}/D$. 

Beyond benchmark evaluation, ARRDE has also shown promising application-level performance in a recent self-consistent ion beam analysis study, where it was highly competitive against several advanced population-based optimizers under a restricted evaluation budget and was selected as the default global optimizer in the AutoNRA framework~\citep{muzakka2026autonra}.

To support reproducibility, all algorithms and CEC benchmark problems used in this work, including ARRDE, are implemented within the Minion framework~\citep{muzakka_2025_14893994}, a C++ and Python library for designing and evaluating optimization algorithms.

\section{Related Work}\label{relwork}

\subsection{LSHADE} 
The Linear Population Size Reduction Success-History-based Adaptive Differential Evolution (L-SHADE) algorithm is an advanced variant of DE that won the CEC2014 competition~\citep{b6900380}. It builds on the foundation of the Success-History-based Adaptive Differential Evolution (SHADE) algorithm~\citep{b6557555}, which ranked fourth in the CEC2013 competition. LSHADE and SHADE share several key features, with the primary difference being L-SHADE's linear population size reduction mechanism, which gradually decreases the population size as iterations progress.

The parameter adaptation process in L-SHADE is built around the historical memory of successful parameter settings. This memory stores past successful values for $F$ and $CR$, and new values are generated by sampling from the memory at each generation. Specifically, let $M_{CR}$ and $M_{F}$ denote the historical memories for $CR$ and $F$, respectively. The memories contain $H$ entries, each storing the $CR$ and $F$ values that led to improvements in previous generations. For each individual in the population, a random index $r_i$ is selected from the range $[1, H]$. Then, $CR$ and $F$ are generated by sampling around the stored values in $M_{CR}$ and $M_{F}$ as follows:

\begin{align}
	CR_i &= 
	\begin{cases} 
		0 & \text{if } M_{CR, r_i} = \perp \\
		\text{randn}(M_{CR, r_i}, 0.1) & \text{otherwise}
	\end{cases}\\
	F_i &= \text{randc}(M_{F, r_i}, 0.1)
\end{align}
where $\text{randn}(m, \sigma)$ represents a normal distribution centered around $m$ with a standard deviation $\sigma$, and $\text{randc}(m, \sigma)$ is a Cauchy distribution used for generating $F$. After mutation and crossover, trial vectors are generated. Once the trial vectors are evaluated, the parameter values that contributed to successful offspring (i.e., vectors that performed better than their parents) are stored in $S_{CR}$ and $S_F$, and these values are used to update the historical memory for the next generation. The historical memory $M_{CR}$ and $M_F$ are updated at the end of each generation using the weighted Lehmer mean, 
where the weights $w_i$ are associated with each value of $F$ and $CR$. They are derived from the fitness improvements $\Delta f_i$ of the trial vectors, calculated as:

\[
w_i = \frac{\Delta f_i}{\sum_{j=1}^{k} \Delta f_j}
\]
where $\Delta f_i = |f(u_i) - f(x_i)|$ is the fitness improvement between the trial vector $u_i$ and the parent vector $x_i$. This way, larger fitness improvements contribute more to the update of $CR$ and $F$.

Linear population size reduction (LPSR) controls the population size $N_G$ as the search progresses. Starting with an initial population size $N_{\text{init}}$, the population size is gradually reduced to a minimum size $N_{\text{min}}$. The reduction follows a linear schedule, which is calculated after each generation as:
\[
N_{G+1} = \left\lceil \left( \frac{N_{\text{min}} - N_{\text{init}}}{\text{MAX\_NFE}} \right) \times \text{NFE} + N_{\text{init}} \right\rceil
\]
Where $\text{NFE}$ is the current number of function evaluations and $\text{MAX\_NFE}$ is the maximum allowed number of evaluations. When $N_{G+1}$ is smaller than $N_G$, the individuals with the worst fitness values are removed.

This gradual reduction in population size allows for extensive exploration in the early phases of the search and increased exploitation as the population size decreases, which helps fine-tune the solution in the later stages.

\subsection{jSO Algorithm}
The jSO algorithm~\citep{7969456} is a DE variant derived from LSHADE, designed to improve its performance, particularly in high-dimensional problems. It ranked second in the CEC2017 competition. It employs a weighted mutation strategy: 
\begin{equation}
	v_i = x_i + F_w \, F (x_{pbest} - x_i) + F (x_{r_1} - x_{r_2}),
\end{equation}
where $F_w$ is a time-dependent weight based on the ratio of function evaluations $\tau = N_{\text{eval}}/N_{\text{max}}$. The value of $F_w$ increases from $0.7$ to $1.2$ as $\tau$ grows. 

In addition, the crossover rate ($CR$), scaling factor ($F$), and weighting factor ($F_w$) are each clamped according to $\tau$ to slow down the early-stage search and promote more refined steps later in the run. The $p$-best parameter, which controls the proportion of elite individuals used in the mutation strategy, is also reduced linearly from $0.25$ to $0.125$ as the search progresses.

Finally, the last entries of the historical memories for $F$ and $CR$ are fixed at $0.9$, ensuring that a subset of individuals always explores more aggressively. These adjustments collectively provide smoother parameter transitions and more stable convergence across different problem scales.

\begin{center}
	\begin{algorithm}
		\caption{Adaptive Restart--Refine Differential Evolution (ARRDE)}
		\label{arrde}
		\begin{algorithmic}[1]
			\State \textbf{Input:} objective $f$, dimension $D$, bounds $[L,U]$, maximum evaluations $N_{\max}$
			\State Compute $N_0$ using~(\ref{N0}) and initialize LSHADE state $(M_F,M_{CR},A)$
			\State Generate initial population $P$ of size $N_0$ by Latin hypercube sampling in $[L,U]$
			\State Evaluate $P$; set $N_{evals}\gets N_0$; set $x^\star\gets \arg\min_{x\in P} f(x)$
			\State Initialize exclusion intervals $\mathcal{E}\gets\emptyset$, final-refinement flag $\mathsf{final}\gets \text{false}$, and consecutive restart count $n_{\mathrm{rest}}\gets 0$
			\While{$N_{evals} < N_{\max}$}
				\State $t \gets N_{evals}/N_{\max}$
				\State Set target population size $N_p(t)$ using~(\ref{eq:np})
				\If{$|P| > N_p(t)$}
					\State Remove the worst $|P|-N_p(t)$ individuals from $P$ and resize $A$
				\EndIf
				\State Initialize success sets $S_F,S_{CR},\Delta f \gets \emptyset$ and generation-improvement flag $\mathsf{improved}\gets \text{false}$
				\For{each individual $x_i \in P$}
					\State Sample $(F_i,CR_i)$ from $(M_F,M_{CR})$ using the LSHADE/jSO rules
					\State Generate trial vector $u_i$ by $current$-to-$p$best mutation and binomial crossover
					\State Repair $u_i$ if needed, evaluate $f(u_i)$, and set $N_{evals}\gets N_{evals}+1$
					\State Apply one-to-one selection and update $P$, $A$, $S_F$, $S_{CR}$, and $\Delta f$
					\If{$f(u_i) < f(x^\star)$}
						\State $x^\star \gets u_i$ and $\mathsf{improved}\gets \text{true}$
					\EndIf
					\If{$N_{evals} \ge N_{\max}$}
						\State \textbf{break}
					\EndIf
				\EndFor
				\If{$S_F \neq \emptyset$}
					\State Update $M_F$ and $M_{CR}$ from $S_F$, $S_{CR}$, and $\Delta f$ using the LSHADE/jSO memory rules
				\EndIf
				\State Compute convergence indicator $s$ from the fitness distribution of $P$
				\If{$t \ge 0.9$}
					\State $\mathsf{final}\gets \text{true}$
				\EndIf
				\If{$s \le s_{\mathrm{tol}}$ \textbf{or} $\mathsf{final}=\text{true}$}
					\State Archive the current population and parameter memories
					\State Update exclusion intervals $\mathcal{E}$ as described in Sec.~\ref{arrde_sec}
					\If{$\mathsf{improved}=\text{false}$ \textbf{and} $n_{\mathrm{rest}} < 1+4t$ \textbf{and} $\mathsf{final}=\text{false}$}
						\State Regenerate $P$ by uniform sampling in $[L,U]$ subject to $\mathcal{E}$ \Comment{restart}
						\State $n_{\mathrm{rest}}\gets n_{\mathrm{rest}}+1$
					\Else
						\State Regenerate $P$ from archived individuals \Comment{refinement}
						\If{$\mathsf{final}=\text{true}$}
							\State Insert $x^\star$ into $P$
						\EndIf
						\State $n_{\mathrm{rest}}\gets 0$
					\EndIf
					\State Evaluate the new population, update $N_{evals}$ accordingly, and refresh $x^\star$
				\EndIf
			\EndWhile
			\State \textbf{return} $x^\star$
		\end{algorithmic}
	\end{algorithm}
\end{center}

\subsection{LSHADE-cnEpSin Algorithm}

The LSHADE-cnEpSin algorithm~\citep{7969336} extends LSHADE by combining sinusoidal adaptation of the scaling factor $F$ with a covariance-matrix-based crossover operator. The sinusoidal mechanism promotes a dynamic balance between exploration and exploitation, while the covariance-based crossover captures variable dependencies through a locally rotated coordinate system. These features make LSHADE-cnEpSin a strong LSHADE-family baseline.

\subsection{NL-SHADE-RSP Algorithm}
The NL-SHADE-RSP algorithm~\citep{9504959} is a variant of the LSHADE-RSP line~\citep{8477977} that achieved first place in the CEC2021 benchmark suite. Its main features are rank-based selective pressure in mutation, adaptive use of the external archive, and multiple crossover operators. These mechanisms make it particularly competitive in low-dimensional settings with large evaluation budgets.

\subsection{NL-SHADE-LBC Algorithm}
The NL-SHADE-LBC algorithm~\citep{nlsade_lbc_2022} is a further development of the NL-SHADE-RSP line proposed for the CEC2022 benchmark suite, where it ranked second. Its main additions are a linear bias change in the adaptation of $F$ and $CR$, nonlinear population size reduction, stronger rank-based selective pressure, and a repeated resampling strategy for bound-constraint handling. The method also modifies archive replacement to favor better stored solutions. These design choices make NL-SHADE-LBC a relevant baseline for low-dimensional problems under intermediate-to-large evaluation budgets, particularly in the CEC2022 regime.

\subsection{j2020 Algorithm}
The j2020 algorithm~\citep{9185551} is a DE variant derived from jDE100~\citep{8789904} and was designed for low-dimensional optimization under very large evaluation budgets. It uses a dual-population architecture, in which a larger population emphasizes global exploration while a smaller population focuses on local refinement. Together with crowding-based selection and separate restart strategies, this design makes j2020 a relevant baseline for exploration-dominated CEC2020-style settings.

\section{The Proposed ARRDE Algorithm}\label{arrde_sec}

ARRDE is built on top of jSO, itself derived from LSHADE, with the goal of improving robustness across different optimization regimes. Starting from jSO, the method is designed to preserve strong performance in high-dimensional moderate-budget settings while improving behavior in low-dimensional large-budget and restricted-budget cases. The proposed method combines three components: an adaptive restart--refine mechanism, a nonlinear, dimension-dependent population reduction schedule, and a budget-aware population-initialization rule. The first two components are the main methodological mechanisms driving the regime-adaptive search behavior, while the initialization rule plays an important supporting role in restricted-budget settings. Although ARRDE introduces additional design parameters compared with the base algorithm, the intention is not to propose a radically new DE framework, but rather a targeted regime-adaptive extension of jSO. Some parameter settings were selected empirically using representative cases from CEC2017 and CEC2020; accordingly, the results on those suites should be interpreted as evidence on the calibration set rather than as fully independent confirmation. Our stronger evidence for transfer therefore comes from CEC2011, CEC2019, and CEC2022, which were not used in tuning these components, as well as from the varying-budget experiments reported later.

Our starting point is jSO, which uses linear population reduction and performs strongly on high-dimensional problems with moderate budgets, such as those in CEC2017, where dimensions up to $D=100$ are considered under the standard budget $N_{\max}=10^4D$. However, algorithms that perform well in this regime often suffer when transferred to the low-dimensional, exploration-oriented settings of CEC2020, where $D\leq 20$ and the evaluation budgets are substantially larger than $10^4D$. In addition, we want the algorithm to remain effective when $N_{\max}$ is very small, which is often the case in practical optimization. ARRDE addresses these difficulties through three components. The first is an adaptive restart--refine mechanism, which includes restart decisions, archive-based refinement, final-stage refinement around the best-so-far solution, and local exclusion regions during restart. The second is a nonlinear, dimension-dependent population reduction schedule that controls how rapidly contraction occurs and, in turn, how strongly restart--refine is activated across regimes. The third is an $N_{\max}$-dependent initial population size used as a practical budget-aware initialization rule.

Apart from the three components described below, ARRDE retains the main jSO framework, including success-history adaptation, the external archive, and progress-dependent parameter control. ARRDE also employs a stochastic boundary-repair strategy: when a candidate violates the search bounds, the infeasible coordinate is re-sampled uniformly from a truncated interval adjacent to the violated boundary, with the interval width limited by the magnitude of the violation. This yields small corrective steps while preserving diversity near the boundary.

The first main component is the restart--refine procedure used after stagnation. To improve performance in low-dimensional, large-budget settings such as CEC2020, ARRDE requires a mechanism that can reintroduce exploration after the population has contracted. The ablation study summarized in Table~\ref{tab:ablation} shows that this mechanism already provides major performance gains relative to the jSO baseline in such regimes, although it slightly reduces performance on CEC2017 when used without the proposed nonlinear reduction. Stagnation is quantified by
\[
s = \operatorname{std}(f(P))/\operatorname{mean}(f(P)).
\]
A restart--refine trigger is activated when $s < s_{\mathrm{tol}}$ or at scheduled points, most notably when $t=0.9$, where
\[
t = N_{\text{evals}} / N_{\max}
\]
denotes normalized search progress. At each trigger, the current converging population, fitness values, and parameter memories are stored in an archive. The subsequent decision between restart and refinement is intended to balance exploration and exploitation. Restart promotes exploration by moving the population to new regions, which is especially useful in low-dimensional, large-budget settings. However, restart alone can weaken intensification around promising areas. Refinement is therefore introduced to recover exploitation by reusing archived search information.

A restart is selected when no improvement has been observed and the number of consecutive restarts has not exceeded
\[
N_{\mathrm{rest},\max} = 1 + 4t;
\]
otherwise, a refinement step is performed. Restart cannot be triggered solely on the absence of improvement in the best-so-far solution, because this would allow the algorithm to keep restarting indefinitely once progress becomes difficult. The threshold $N_{\mathrm{rest},\max}=1+4t$ therefore imposes a restart limit that increases from 1 at the beginning of the run to 5 near termination, making the procedure progressively more explorative as the search advances.

In a refinement step, the population is not reinitialized randomly; instead, individuals are sampled at random from the archived converging populations collected at previous triggers. As a result, the best-so-far solution is not necessarily included in every refinement phase. This choice preserves additional exploration while the search remains focused on exploiting previously promising regions. During the final refinement stage, however, the best-so-far solution is explicitly reinserted into the population to strengthen late-stage intensification.

Local exclusion during restart is part of the restart--refine mechanism. Its purpose is to improve exploration in cases where the search is sensitive to the initial population or restart location, so that repeatedly sampling near previously visited regions becomes undesirable. In practice, its impact is most relevant when $N_{\max}$ is very large and the algorithm restarts multiple times, as in CEC2020 and CEC2019. For each dimension $d$, exclusion intervals
\[
[\ell_d, u_d] = [\max(\mu_d - \sigma_d, L_d), \min(\mu_d + \sigma_d, U_d)]
\]
are constructed from archived populations and merged across restarts. Newly generated individuals are then resampled outside these intervals. This discourages the restarted population from revisiting regions that have already been explored repeatedly, thereby improving diversity and making the restart stage less dependent on favorable initialization.

The second main component is the nonlinear, dimension-dependent population-size schedule. Its role is to determine how quickly the population contracts during the main search phase and, as a consequence, how easily the restart--refine mechanism is activated in different regimes. Because this component is a central regime-adaptive element of ARRDE, its effect is examined separately in the ablation study summarized in Table~\ref{tab:ablation}. The ablation shows that, after the substantial gains already provided by restart--refine in low-dimensional settings, the proposed nonlinear schedule further improves performance on CEC2020 and CEC2022 while also bringing the CEC2017 results back closer to the strong jSO baseline. At $t=0.9$, when final refinement is enforced, the population is intentionally expanded and then reduced again toward its terminal size. The resulting piecewise schedule is defined as follows:
\begin{equation}\label{eq:np}
	N_p(t) =
	\begin{cases}
		\max\!\left(4,\; N_{0} - \left(N_{0}-\frac{D}{2}\right)\!\left[1 - \left(\frac{0.9-t}{0.9}\right)^{r}\right]\right), & t \le 0.9, \\[0.8ex]
		\max\!\left(4,\; \frac{N_0}{4} - \left(\frac{N_0}{4} - \frac{D}{2}\right)\!\left[1 - \left(\frac{1 - t}{0.1}\right)^2\right]\right), & t > 0.9 .
	\end{cases}
\end{equation}
The decay exponent
\begin{equation}\label{rr}
	r = 1.17 + 2.075\, e^{-0.0567D}
\end{equation}
controls the strength of the contraction in the main search phase and decreases with increasing dimensionality. The first branch governs the main search phase and contracts the population toward approximately $D/2$ by $t=0.9$. The second branch corresponds to the final-refinement stage: the population is first re-expanded through archive-based reinitialization and is then reduced again, from that enlarged state, toward the final target size. Thus, the second branch should not be interpreted as a continuation of monotone contraction from the first branch, but as the scheduled population evolution within final refinement. Note here that the functional form in~\eqref{rr} and its parameters were selected empirically on representative benchmark cases from CEC2017 and CEC2020. Whereas jSO reduces the population down to a minimum of 4, ARRDE uses 4 only as a hard lower bound and otherwise follows a larger $D/2$ target, because a population of 4 was found insufficient to maintain diversity for effective late-stage refinement.

This design changes the search behavior systematically across regimes. In low-dimensional problems, the stronger contraction accelerates the onset of stagnation and therefore activates restart--refine earlier and more often, which is advantageous over long evaluation horizons such as those of CEC2020. In high-dimensional problems, the schedule remains much closer to the approximately linear jSO behavior, so the search dynamics stay near a baseline that is already strong on CEC2017. This regime dependence is intentional. Restart is more useful when $D$ is small, because the search can be redirected to new regions without incurring a severe loss of coverage. In large dimensions, however, restart is less useful because the search space expands rapidly, and maintaining a larger population is generally more important for efficient exploration. 

The third component is the budget-aware initialization rule, which sets the initial population size. Since very small budgets cannot support an unnecessarily large starting population, ARRDE scales the initial population with the available budget. The population is generated using Latin hypercube sampling within the variable bounds, and its size is defined as
\begin{align}
	N_{0} &= D \times \max\!\left[2,\; 2+ 5.756(\eta - 2)^{1.609}\right],\label{N0} \\
	\eta  &= \log_{10}(N_{\max}/D),
\end{align}
where $\eta=\log_{10}(N_{\max}/D)$ measures the normalized evaluation budget. This rule makes $N_0$ increase when more evaluations are available, while keeping the initial population smaller when the budget is limited. This is particularly relevant in practical small-budget scenarios, where spending too many evaluations on a large initial population can be wasteful. At the same time, when $N_{\max}=10^4D$, the resulting population size remains close to the standard jSO-style regime, so this component behaves as a natural extension of the baseline rather than a radical departure from it. The coefficients were determined empirically using the CEC2017 and CEC2020 benchmark suites.

The complete ARRDE procedure is summarized in Algorithm~\ref{arrde}.

\section{Numerical Experiments and Discussions}

\subsection{Experimental Setup}

To evaluate the performance of the proposed ARRDE algorithm, we conducted extensive numerical experiments on benchmark suites from CEC2017~\citep{cec2017}, CEC2020~\citep{cec2020}, CEC2022~\citep{cec2022}, the 100-Digit Challenge from CEC2019~\citep{cec2019}, and real-world optimization problems from CEC2011~\citep{cec2011}. For all benchmark suites, the maximum number of function evaluations ($N_{\max}$) was set according to the respective competition rules. Each experiment was repeated 51 times to ensure statistical reliability. The run index was used as the seed for the global random number generator to guarantee reproducibility. All algorithms and test functions were implemented in C++, compiled using MSVC, and executed on a Windows~11 workstation.

The CEC2017, CEC2020, and CEC2022 benchmark suites consist of four categories of problems: unimodal, basic, hybrid, and composition functions. The problems are typically shifted, rotated, and include bias terms in their final objective values. Unimodal functions, such as the Bent Cigar and Zakharov functions, are single-peak problems that evaluate search efficiency. Basic functions include well-known multimodal landscapes such as Rastrigin, Schaffer, Lévy, and Schwefel functions. Hybrid functions combine the values of $N$ basic functions computed over randomly assigned subcomponents of the search space. Composition functions are the most challenging category, formed by weighted mixtures of $N$ basic functions where the weight distribution depends on the decision vector. These weight functions ensure that the global optimum of a composition function coincides with the optimum of one of its constituent basic functions. In the CEC2017 suite, there are 3 unimodal, 7 basic, 10 hybrid, and 10 composition functions. CEC2020 contains 1 unimodal, 3 basic, 3 hybrid, and 3 composition functions, while CEC2022 includes 1 unimodal, 4 basic, 3 hybrid, and 4 composition functions. 

For CEC2017, the algorithm was tested on 29 problems at dimensions 10, 30, 50, and 100, with the maximum number of function evaluations set to $N_{\max} = 10^4 \times D$. Following the CEC2017 guidelines, problem F2 (the shifted and rotated Rastrigin function) was excluded due to numerical instability in higher dimensions. For CEC2020, the algorithm was evaluated on 10 problems at dimensions 5, 10, 15, and 20, with the corresponding evaluation budgets set to $N_{\max} = 5\times10^4$, $10^6$, $3\times10^6$, and $10^7$. For CEC2022, the algorithm was tested on 12 problems at dimensions 10 and 20, using $N_{\max} = 2\times10^5$ and $10^6$, respectively. It is worth noting that although the CEC2017 suite contains higher-dimensional problems, it uses substantially lower evaluation budgets compared with the CEC2020 and CEC2022 suites. Conversely, CEC2020 represents the opposite extreme: relatively low-dimensional problems paired with exceptionally large evaluation budgets.

For the CEC2019 100-Digit Challenge, algorithms are evaluated under an effectively unlimited time budget. In this study, we impose a practical limit of $N_{\max} = 10^8$. The dimensionality of the problems ranges from 9 to 18, with most being 10-dimensional. For each problem, the number of correctly retrieved digits (up to the 10th decimal place) is recorded, and the final ranking is determined based on the average number of correct digits achieved in the best 25 out of 51 runs.

The CEC2011 real-world optimization suite comprises 22 problems with dimensionalities ranging from 6 to 212. These problems are derived from simplified formulations of practical engineering tasks, including FM sound wave parameter estimation, Lennard--Jones and Tersoff potential minimization, spread-spectrum radar polyphase code design, transmission network expansion planning (TNEP), transmission pricing, circular antenna array design, static and dynamic economic load dispatch (ELD/DED), hydrothermal scheduling, and spacecraft trajectory optimization for the \emph{Messenger} and \emph{Cassini~2} missions. Several of the original problems include inequality constraints in addition to bound constraints. Since our focus in this study is on bound-constrained optimization, these inequality constraints are omitted. Despite their simplifications, many CEC2011 problems remain very challenging due to their high dimensionality and multimodal landscapes. Following the CEC2011 benchmarking protocol, we evaluate all algorithms under three function-evaluation budgets: $N_{\max} = 5\times10^4$, $10^5$, and $1.5\times10^5$.

These benchmark suites collectively assess an algorithm's ability to address a broad spectrum of optimization challenges, including separability, multimodality, rotation, nonlinearity, and real-world modeling complexity. The experimental results and comparisons with state-of-the-art algorithms are presented and analyzed in the subsequent sections.

In the numerical experiments, ARRDE is compared with six representative state-of-the-art DE variants reviewed in Section~\ref{relwork}: LSHADE, jSO, LSHADE-cnEpSin, j2020, NL-SHADE-RSP, and NL-SHADE-LBC. These methods were selected to cover strong DE baselines from the benchmark regimes most relevant to this study. jSO and LSHADE-cnEpSin ranked second and third, respectively, in CEC2017 and therefore represent strong high-dimensional, moderate-budget baselines. NL-SHADE-RSP ranked first in CEC2021 and also showed stronger performance than j2020 on CEC2020 in our tests, while j2020 ranked third in CEC2020 and serves as a relevant low-dimensional, large-budget baseline. NL-SHADE-LBC ranked second in CEC2022 and is included as an additional recent baseline that is particularly relevant to the CEC2022 benchmark regime. The selected baselines were those with stable and reproducible implementations that could be integrated into a unified evaluation framework; although additional recent competition winners exist, not all were available in a form that allowed reliable like-for-like comparison within the present study. All algorithms use the parameter settings recommended in their original publications. The presentation in this section therefore focuses on aggregated performance using the scoring metrics described below.

\subsection{Scoring Methodology}

To obtain a balanced and reliable assessment, both \emph{rank-based} and \emph{accuracy-based} performance metrics are employed. The official metrics of the considered CEC benchmark suites are still reported in the corresponding result tables where applicable. However, because those official metrics emphasize different performance aspects and are therefore not directly comparable across suites, we additionally define a unified robustness-oriented metric for cross-suite assessment. Its role is to provide a common aggregation when the objective is not to reproduce the exact ranking protocol of any single CEC suite, but to compare overall behavior across heterogeneous suites under a consistent error normalization and rank aggregation. Let $k$ index the algorithm, $j$ the benchmark problem, $i$ the independent run, $D$ the problem dimensionality, $N_p$ the number of benchmark problems, and $N_{\text{runs}}$ the number of independent runs.

For rank-based evaluation, we adopt the Friedman ranking~\citep{Friedman01121937}. For each problem and run, algorithms are ranked according to their final objective function values, with lower values indicating better performance (ties receive the average rank). Denoting by $r_{i,j,k}$ the rank of algorithm $k$ on problem $j$ in run $i$, the overall Friedman rank-based score is defined as
\begin{equation}\label{Rk}
	R_k
	= \frac{1}{N_p\,N_{\text{runs}}}
	\sum_{j=1}^{N_p} \sum_{i=1}^{N_{\text{runs}}} r_{i,j,k}.
\end{equation}
Smaller values of $R_k$ indicate superior performance. For pairwise win/tie/loss (W/T/L) comparisons between algorithms, statistical significance is assessed using the Mann--Whitney~U test~\citep{10.1214/aoms/1177730491} with a significance level of $\alpha=0.05$.

Accuracy-based performance is quantified using the mean absolute relative error. For algorithm $k$ and problem $j$, it is defined as
\begin{equation}
	\epsilon_{k,j}
	= \frac{1}{N_{\text{runs}}}
	\sum_{i=1}^{N_{\text{runs}}}
	\left|
	\frac{f_j(x_{i,k,j}) - f_j(x_j^*)}{f_j(x_j^*)}
	\right|,
\end{equation}
where $x_{i,k,j}$ denotes the solution obtained by algorithm $k$ on problem $j$ in run $i$, $f_j(\cdot)$ is the objective function of problem $j$, and $x_j^*$ is the known global optimum. In the implementation, if $f_j(x_j^*)=0$, the denominator is replaced by $1$ to avoid division by zero; this occurs only once in the present study, for CEC2011 F1. For CEC2011, where the true optimum is unavailable, $x_j^*$ is approximated by the best solution found across all algorithms and runs.

To prevent domination by outlier problems, the relative error is mapped to a bounded interval using
\begin{equation}
	\mathcal{E}_{k,j} = \frac{\epsilon_{k,j}}{1 + \epsilon_{k,j}},
\end{equation}
which maps errors to $[0,1)$. This transformation preserves fine-grained accuracy differences for small errors, while ensuring that large errors saturate and cannot dominate the aggregated score. By bounding each problem's contribution, all benchmark functions contribute comparably to the overall evaluation, regardless of scale or difficulty. The overall accuracy-based score of algorithm $k$ is obtained by averaging $\mathcal{E}_{k,j}$ over all problems for the given dimension.

Following the cross-dimension aggregation scheme used in CEC2017 and CEC2020, accuracy-based and rank-based scores are combined across dimensions. Let $\mathcal{E}_k^{(D)}$ and $R_k^{(D)}$ denote the accuracy-based and rank-based scores of algorithm $k$ at dimension $D$, respectively. Using the dimension-dependent weights $w_D$, we define
\begin{align}
	S_{E,k} &= \sum_D w_D \, \mathcal{E}_k^{(D)}, \\
	S_{R,k} &= \sum_D w_D \, R_k^{(D)},
\end{align}
where $w_D=\{0.1,0.2,0.3,0.4\}$ for increasing dimensions in CEC2017 ($D=10,30,50,100$) and CEC2020 ($D=5,10,15,20$). For CEC2022 ($D=10,20$), we follow the same CEC2017-style weighting scheme and use $w_D=\{0.1,0.2\}$. For CEC2011, where no cross-dimension aggregation is required, we set $w_D=1$.

The overall combined score for algorithm $k$ is then given by
\begin{equation}\label{Stot}
	S_{\text{tot},k}
	= 50 \left(
	\frac{\min_l S_{E,l}}{S_{E,k}}
	+
	\frac{\min_l S_{R,l}}{S_{R,k}}
	\right),
\end{equation}
where $l$ indexes the competing algorithms. This unified score is used as an additional indicator for cross-suite robustness assessment; it is not intended to replace the official metrics of the individual benchmark suites.

Similarly, to facilitate per-dimension analysis, we also define a dimension-specific combined score. For a given dimension $D$, the combined score of algorithm $k$ is defined as
\begin{equation}
	S_k^{(D)}
	= 50 \left(
	\frac{\min_l \mathcal{E}_l^{(D)}}{\mathcal{E}_k^{(D)}}
	+
	\frac{\min_l R_l^{(D)}}{R_k^{(D)}}
	\right),
\end{equation}
 This per-dimension score enables direct comparison of algorithm performance at a fixed dimensionality, complementing the aggregated score $S_{\text{tot},k}$.

For completeness, we also include the original scoring procedures used in CEC2017 and CEC2020. In CEC2017, the overall scoring formula is structurally similar to (\ref{Stot}), but the rank-based and accuracy-based components differ. In the original CEC2017 and CEC2020 rules, the rank-based component $R_k$ is computed from the ranks of the \emph{mean} performance values across runs, rather than from Friedman rank. For the accuracy-based component, CEC2017 uses the absolute error
\begin{equation}
	\mathcal{E}^{\mathrm{CEC2017}}_k
	= \sum_j^{N_p} \sum_{i=1}^{N_{\text{runs}}}
	\left( f_j(x_i) - f_j(x^*_j) \right),
\end{equation}
where \(i\), \(j\), and \(k\) index the run, problem, and algorithm, respectively. In CEC2020, the accuracy score is defined using a normalized error:
\begin{equation}
	\mathcal{E}^{\mathrm{CEC2020}}_k
	= \sum_j^{N_p}
	\frac{f_j(x^{\text{best}}_{j,k}) - f_j(x^*_j)}
	{f_j(x^{\text{best,max}}_j) - f_j(x^*_j)},
\end{equation}
where \(f_j(x^{\text{best}}_{j,k})\) is the best value obtained by algorithm \(k\), and \(f_j(x^{\text{best,max}})\) is the worst among the best values obtained across all algorithms.

For the CEC2019 100-Digit Challenge, in addition to reporting the $\mathcal{E}$ and $S$ scores described earlier, we also include its official scoring procedure. In this scheme, algorithms are ranked according to the average number of correctly recovered digits, computed from their best 25 runs. 

While these official formulations are appropriate for their respective competitions, they are less suitable for the robustness-oriented assessment considered here. The CEC2017 score aggregates absolute errors and can therefore be dominated by problems with large objective-value scales, whereas the CEC2020 score is based only on the single best result of each algorithm and is sensitive to the relative spread among competitors on each problem. Both choices are reasonable for competition ranking, but they are less aligned with the present goal of assessing typical performance across heterogeneous benchmarks. For this reason, the proposed ARRDE is fine-tuned and evaluated primarily using the robustness-oriented score $S_{\text{tot}}$ in (\ref{Stot}), which is based on bounded relative error and Friedman ranking.

\subsection{Results}

\subsubsection{Ablation Study}

Table~\ref{tab:ablation} summarizes a compact ablation study of the staged ARRDE design using four variants: the original \textit{jSO}, \textit{jSO-RR-LR} with the proposed restart--refine mechanism while retaining linear reduction, \textit{jSO-RR-NLR} with the proposed nonlinear reduction, and the full \textit{ARRDE} with the additional $N_{\max}$-dependent initialization. Several patterns are clear. First, adding restart--refine already yields major gains on the low-dimensional, large-budget CEC2020 and CEC2022 suites, confirming that this mechanism is one of the two main methodological components of ARRDE, although it slightly reduces performance on CEC2017 when used together with linear reduction. Second, replacing linear reduction by the proposed nonlinear, dimension-dependent schedule improves the low-dimensional suites further while also bringing the high-dimensional CEC2017 behavior back closer to the strong jSO baseline. Finally, the full ARRDE variant provides the strongest overall robustness, indicating that the budget-dependent initialization gives an additional practical gain after the two main methodological components are in place. Taken together, the ablation results suggest that the proposed components are not acting merely as a zero-sum benchmark-specific adjustment. Instead, their combination produces a more uniformly improved cross-suite profile, which is consistent with the interpretation of ARRDE as a more robust variant. Nevertheless, because some components were calibrated on CEC2017 and CEC2020, this evidence should be read as suggestive rather than definitive, and the transfer results on CEC2011, CEC2019, and CEC2022 are therefore particularly important. 

\begin{table*}[t]
	\centering
	\caption{Compact ablation study using the combined score $S$ (higher is better). RR denotes restart--refine, ER exclusion region, LR linear reduction, and NLR the proposed nonlinear dimension-dependent reduction.}
\label{tab:ablation}
		\TableFont
	\begin{tabular}{l|ccccc|ccccc|ccc}
		\hline
		Algorithm & \multicolumn{5}{c|}{CEC2017} & \multicolumn{5}{c|}{CEC2020} & \multicolumn{3}{c}{CEC2022} \\
		& 10D & 30D & 50D & 100D & $S_{\text{tot}}$ & 5D & 10D & 15D & 20D & $S_{\text{tot}}$ & 10D & 20D & $S_{\text{tot}}$ \\
		\hline
		jSO & 90.438 & \textbf{97.654} & \textbf{98.504} & 97.178 & \textbf{99.223} & 63.095 & 54.642 & 55.929 & 65.964 & 60.874 & 86.198 & 61.189 & 66.496 \\
		jSO-RR-LR & 95.210 & 97.553 & 95.063 & 93.356 & 96.668 & 70.950 & 70.775 & 73.351 & 85.952 & 78.781 & 87.948 & 77.963 & 80.816 \\
		jSO-RR-NLR & 97.355 & 97.314 & 97.950 & 94.233 & 97.791 & 89.519 & 91.742 & 87.919 & 89.429 & 89.889 & 90.684 & 93.924 & 92.927 \\
		ARRDE & \textbf{100.000} & 92.270 & 95.785 & \textbf{100.000} & 99.218 & \textbf{100.000} & \textbf{100.000} & \textbf{100.000} & \textbf{100.000} & \textbf{100.000} & \textbf{100.000} & \textbf{100.000} & \textbf{100.000} \\
	\hline
	\end{tabular}
\end{table*}

\subsubsection{CEC2017}

On CEC2017, Tables~\ref{2017-10-30}--\ref{2017-50-100} reveal a clear three-tier structure. ARRDE, jSO, and LSHADE-cnEpSin form a tightly grouped top tier, with ARRDE achieving the best overall combined score, followed by jSO and LSHADE-cnEpSin. This ordering mirrors the official CEC2017 competition results, where jSO and LSHADE-cnEpSin ranked second and third. Dimension-wise, ARRDE dominates at 10D and remains extremely close to jSO at 30D and 50D, where jSO attains slightly higher $S$ values. At 100D, LSHADE-cnEpSin takes the lead, with ARRDE a close second. Because the numerical differences among these three algorithms are small, their performance on CEC2017 should be viewed as broadly comparable rather than sharply separated. Indeed, we observed that minor adjustments to ARRDE's parameters can shift the dimension-wise ordering relative to jSO and LSHADE-cnEpSin, which is unsurprising given that ARRDE incorporates several of jSO's design components. 

The second tier is represented by LSHADE and NL-SHADE-LBC, whose performance gap relative to the top group is substantial and consistent across all dimensions.

The third tier consists of j2020 and NL-SHADE-RSP, which perform markedly worse on CEC2017. NL-SHADE-RSP remains competitive at $D=10$, where it ranks third behind ARRDE and jSO, but its performance degrades steadily at higher dimensions, placing sixth at $D=30$, $50$, and $100$. This behavior is expected: both NL-SHADE-RSP and j2020 were developed and tuned primarily for the CEC2020 and CEC2021 suites, which consist of low-dimensional problems ($D \le 20$) with very generous evaluation budgets. Their advantages in those settings do not generalize well to the higher-dimensional, stricter-budget environment of CEC2017.

Finally, we note a small discrepancy between the aggregated combined score $S_{\text{tot}}$ and the official CEC2017 score $S_{2017}$. As discussed earlier, this difference stems from the underlying error-normalization schemes: the official score uses absolute final errors, whereas our accuracy-based measure $\mathcal{E}$ employs a relative-error formulation. Nevertheless, under both scoring methods, ARRDE ranks first overall.

\subsubsection{CEC2020}\label{sec:20}

The CEC2020 results in Tables~\ref{2020-5-10}--\ref{2020-15-20} reveal a markedly different, yet complementary, performance landscape. Although the CEC2020 problems are mostly low-dimensional, their extremely large evaluation budgets promote extensive exploration. Under our combined metric, ARRDE clearly wins overall: it attains the highest $S_{\text{tot}}$ and ranks first in the $S$--score for all four dimensions (5D, 10D, 15D, and 20D). Interestingly, however, the rank-based score $R$ tells a different story---ARRDE does not rank first in any single dimension. NL-SHADE-RSP most frequently achieves the best $R$--score, while j2020 takes second place at 10D. Yet ARRDE's consistently superior $\mathcal{E}$ values compensate for these rank differences, yielding the highest combined performance.

The performance of the algorithms that excel on CEC2017 is also noteworthy. Their ranking is almost completely reversed on CEC2020: LSHADE now outperforms jSO and LSHADE-cnEpSin, and jSO surpasses LSHADE-cnEpSin. Apart from ARRDE, which consistently retains first place in $S_{\text{tot}}$, the ranking structure in CEC2020 is largely anti-correlated with that observed in CEC2017. This further illustrates the strong suite-specific behavior of many DE variants and underscores the robustness advantage exhibited by ARRDE.

We also observe that the official CEC2020 scoring produces rankings that differ substantially from those obtained using our $\mathcal{E}$- and $R$-based metrics. This difference is largely due to the design of the official score $S_{2020}$, where each function error is normalized using the worst of the best results among all compared algorithms. Under this scheme, even very small numerical differences on easy functions can be amplified in the normalized score. A second contributing factor is that $S_{2020}$ is based on the \emph{best} run out of 51, rather than on an average across runs. This choice is natural for competition ranking, where emphasis is placed on the best obtained result, but it does not fully reflect typical performance across repeated trials. Consequently, algorithms such as j2020 and NL-SHADE-RSP can rank above ARRDE under $S_{2020}$ even when ARRDE shows stronger average behavior under the robustness-oriented scoring used here.

	\begin{table*}[!t]
		\centering
		\caption{CEC2017 results (10D, 30D): accuracy $\mathcal{E}$, rank $R$, combined score $S$, and win/tie/loss (W/T/L) of ARRDE against each algorithm over 29 functions. Parentheses indicate ranks; bold denotes best.}

		\label{2017-10-30}
		\TableFont
		\begin{tabular}{l|cccc|cccc}
\hline
Algorithm & \multicolumn{4}{|c|}{10D} & \multicolumn{4}{|c}{30D} \\
 & $\mathcal{E}$ & $R$ & $S$ & W/T/L & $\mathcal{E}$ & $R$ & $S$ & W/T/L \\
\hline
ARRDE & \textbf{0.031 (1)} & \textbf{3.439 (1)} & \textbf{100.000 (1)} & 0/29/0 & 0.094 (2) & 2.968 (2) & 95.740 (2) & 0/29/0 \\
LSHADE & 0.037 (4) & 3.916 (3) & 85.695 (4) & 11/12/6 & 0.105 (4) & 3.277 (4) & 85.934 (4) & 14/6/9 \\
jSO & 0.036 (3) & 3.674 (2) & 89.927 (2) & 8/13/8 & \textbf{0.092 (1)} & \textbf{2.770 (1)} & \textbf{100.000 (1)} & 7/15/7 \\
LSHADE-cnEpSin & 0.069 (7) & 3.998 (4) & 65.456 (7) & 13/11/5 & 0.096 (3) & 3.267 (3) & 90.346 (3) & 13/13/3 \\
j2020 & 0.037 (5) & 4.758 (7) & 77.658 (6) & 19/8/2 & 0.200 (7) & 6.197 (7) & 45.288 (7) & 25/4/0 \\
NL-SHADE-RSP & 0.035 (2) & 4.058 (5) & 87.165 (3) & 12/9/8 & 0.168 (6) & 5.265 (6) & 53.665 (6) & 22/4/3 \\
NL-SHADE-LBC & 0.039 (6) & 4.157 (6) & 81.533 (5) & 14/8/7 & 0.110 (5) & 4.256 (5) & 74.338 (5) & 21/6/2 \\
\hline
	\end{tabular}
	\end{table*}

\begin{table*}[!t]
	\centering
	\caption{CEC2017 results (50D, 100D) with combined cross-dimension scores: accuracy $\mathcal{E}$, rank $R$, combined score $S$, and ARRDE win/tie/loss (W/T/L) over 29 functions. $S_{2017}$ denotes the original CEC2017 score. Parentheses indicate ranks; bold denotes best.}

	\label{2017-50-100}
	\TableFont
\begin{tabular}{l|cccc|cccc|cc}
\hline
Algorithm & \multicolumn{4}{|c|}{50D} & \multicolumn{4}{|c|}{100D} & \multicolumn{2}{|c}{Combined} \\
 & $\mathcal{E}$ & $R$ & $S$ & W/T/L & $\mathcal{E}$ & $R$ & $S$ & W/T/L & $S_{\text{tot}}$ & $S_{2017}$ \\
\hline
ARRDE & 0.160 (2) & \textbf{2.571 (1)} & 99.163 (2) & 0/29/0 & 0.252 (2) & 2.445 (2) & 98.536 (2) & 0/29/0 & \textbf{100.000 (1)} & \textbf{98.984 (1)} \\
LSHADE & 0.169 (4) & 3.349 (4) & 84.937 (4) & 14/9/6 & 0.294 (4) & 3.669 (4) & 74.967 (4) & 23/6/0 & 82.384 (4) &  85.144 (4)  \\
jSO & \textbf{0.157 (1)} & 2.587 (2) & \textbf{99.680 (1)} & 9/11/9 & 0.260 (3) & 2.715 (3) & 92.038 (3) & 11/9/9 & 97.500 (2) &   98.260 (2)\\
LSHADE-cnEpSin & 0.164 (3) & 3.001 (3) & 90.698 (3) & 15/9/5 & \textbf{0.245 (1)} & \textbf{2.442 (1)} & \textbf{100.000 (1)} & 9/7/13 & 95.062 (3) &  90.911 (3) \\
j2020 & 0.403 (7) & 6.239 (7) & 40.081 (7) & 27/2/0 & 0.596 (7) & 6.277 (7) & 40.006 (7) & 27/0/2 & 43.186 (7) &  22.480 (7)\\
NL-SHADE-RSP & 0.362 (6) & 6.005 (6) & 43.083 (6) & 27/2/0 & 0.546 (6) & 6.041 (6) & 42.666 (6) & 28/1/0 & 47.098 (6) &  37.581 (6) \\
NL-SHADE-LBC & 0.184 (5) & 4.247 (5) & 72.811 (5) & 23/4/2 & 0.318 (5) & 4.412 (5) & 66.205 (5) & 24/3/2 & 72.125 (5) &  68.066 (5)\\
\hline
\end{tabular}
\end{table*}


	\begin{table*}[!t]
		\centering
		\caption{CEC2020 results (5D, 10D): accuracy $\mathcal{E}$, rank $R$, combined score $S$, and ARRDE win/tie/loss (W/T/L) over 10 functions. Parentheses indicate ranks; bold denotes best.}

	\label{2020-5-10}
		\TableFont
	\begin{tabular}{l|cccc|cccc}
		\hline
		Algorithm & \multicolumn{4}{|c|}{5D} & \multicolumn{4}{|c}{10D} \\
		& $\mathcal{E}$ & $R$ & $S$ & W/T/L & $\mathcal{E}$ & $R$ & $S$ & W/T/L \\
		\hline
		ARRDE & \textbf{0.008 (1)} & 3.626 (2) & \textbf{91.984 (1)} & 0/10/0 & \textbf{0.009 (1)} & 3.420 (3) & \textbf{93.693 (1)} & 0/10/0 \\
		LSHADE & 0.017 (4) & 4.201 (4) & 59.153 (4) & 2/7/1 & 0.030 (5) & 4.776 (6) & 46.488 (6) & 6/3/1 \\
		jSO & 0.018 (7) & 4.456 (7) & 55.742 (7) & 4/5/1 & 0.029 (4) & 4.284 (5) & 50.960 (5) & 6/1/3 \\
		LSHADE-cnEpSin & 0.018 (6) & 4.282 (6) & 57.357 (6) & 4/5/1 & 0.032 (7) & 5.313 (7) & 42.380 (7) & 7/2/1 \\
		j2020 & 0.010 (2) & 4.120 (3) & 77.840 (3) & 4/6/0 & 0.011 (2) & 3.325 (2) & 87.849 (2) & 4/2/4 \\
		NL-SHADE-RSP & 0.010 (3) & \textbf{3.045 (1)} & 87.442 (2) & 1/5/4 & 0.018 (3) & \textbf{2.988 (1)} & 75.426 (3) & 2/3/5 \\
		NL-SHADE-LBC & 0.017 (5) & 4.270 (5) & 57.881 (5) & 3/6/1 & 0.030 (6) & 3.893 (4) & 53.526 (4) & 4/2/4 \\
		\hline
	\end{tabular}
	\end{table*}

\begin{table*}[!t]
	\centering
	\caption{CEC2020 results (15D, 20D) with combined cross-dimension scores: accuracy $\mathcal{E}$, rank $R$, combined score $S$, and ARRDE win/tie/loss (W/T/L) over 10 functions. $S_{2020}$ denotes the original CEC2020 score. Parentheses indicate ranks; bold denotes best.}

	\label{2020-15-20}
		\TableFont
		\begin{tabular}{l|cccc|cccc|cc}
			\hline
			Algorithm & \multicolumn{4}{|c|}{15D} & \multicolumn{4}{|c|}{20D} & \multicolumn{2}{|c}{Combined} \\
			& $\mathcal{E}$ & $R$ & $S$ & W/T/L & $\mathcal{E}$ & $R$ & $S$ & W/T/L & $S_{\text{tot}}$ & $S_{2020}$ \\
			\hline
			ARRDE & \textbf{0.018 (1)} & 3.278 (2) & \textbf{93.974 (1)} & 0/10/0 & \textbf{0.023 (1)} & 3.257 (2) & \textbf{97.140 (1)} & 0/10/0 & \textbf{94.937 (1)} & 72.032 (3) \\
			LSHADE & 0.034 (4) & 4.161 (4) & 61.489 (4) & 6/2/2 & 0.037 (4) & 4.381 (5) & 66.562 (5) & 7/1/2 & 60.840 (4) &   45.807 (4) \\
			jSO & 0.039 (5) & 4.701 (6) & 54.410 (6) & 7/2/1 & 0.036 (3) & 4.449 (6) & 67.065 (4) & 7/1/2 & 59.445 (5) &  43.987 (5) \\
			LSHADE-cnEpSin & 0.040 (6) & 4.862 (7) & 52.650 (7) & 6/2/2 & 0.044 (7) & 4.959 (7) & 57.759 (7) & 7/1/2 & 53.571 (7) &  38.839 (7)  \\
			j2020 & 0.027 (3) & 3.666 (3) & 73.492 (3) & 5/2/3 & 0.039 (5) & 3.610 (3) & 72.473 (3) & 4/1/5 & 74.001 (3) & 90.397 (2)  \\
			NL-SHADE-RSP & 0.022 (2) & \textbf{2.883 (1)} & 91.491 (2) & 2/2/6 & 0.032 (2) & \textbf{3.071 (1)} & 86.696 (2) & 4/2/4 & 86.360 (2) & \textbf{98.926 (1)} \\
			NL-SHADE-LBC & 0.040 (7) & 4.449 (5) & 55.275 (5) & 5/4/1 & 0.042 (6) & 4.274 (4) & 63.580 (6) & 4/5/1 & 59.012 (6) &  43.174 (6) \\
			\hline
	\end{tabular}
		\end{table*}


\begin{table*}
	\centering
	\caption{CEC2022 results (10D, 20D) with combined cross-dimension scores: accuracy $\mathcal{E}$, rank $R$, combined score $S$, and ARRDE win/tie/loss (W/T/L) over 12 functions. Parentheses indicate ranks; bold denotes best.}		
	\label{2022}
		\TableFont
		\begin{tabular}{l|cccc|cccc|c}
		\hline
		Algorithm & \multicolumn{4}{|c|}{10D} & \multicolumn{4}{|c|}{20D} & \multicolumn{1}{|c}{Combined} \\
		& $\mathcal{E}$ & $R$ & $S$ & W/T/L & $\mathcal{E}$ & $R$ & $S$ & W/T/L & $S_{\text{tot}}$  \\
		\hline
		ARRDE & 0.014 (3) & \textbf{3.497 (1)} & 94.196 (2) & 0/12/0 & \textbf{0.017 (1)} & \textbf{2.774 (1)} & \textbf{100.000 (1)} & 0/12/0 & \textbf{100.000 (1)}  \\
		LSHADE & 0.017 (6) & 4.609 (7) & 75.869 (7) & 6/6/0 & 0.035 (4) & 4.097 (5) & 58.060 (6) & 6/6/0 & 63.090 (6) \\
		jSO & 0.016 (5) & 4.093 (5) & 81.959 (5) & 5/6/1 & 0.035 (3) & 3.830 (2) & 60.557 (3) & 6/6/0 & 66.550 (3)  \\
		LSHADE-cnEpSin & 0.017 (7) & 4.369 (6) & 76.006 (6) & 5/7/0 & 0.035 (6) & 4.070 (4) & 58.153 (5) & 5/7/0 & 63.516 (5)  \\
		j2020 & \textbf{0.013 (1)} & 3.773 (3) & \textbf{96.340 (1)} & 2/9/1 & 0.026 (2) & 4.481 (6) & 63.940 (2) & 7/4/1 & 73.107 (2)  \\
		NL-SHADE-RSP & 0.014 (2) & 3.893 (4) & 90.734 (3) & 3/8/1 & 0.038 (7) & 4.732 (7) & 51.590 (7) & 8/3/1 & 60.626 (7)   \\
		NL-SHADE-LBC & 0.016 (4) & 3.766 (2) & 85.949 (4) & 3/8/1 & 0.035 (5) & 4.016 (3) & 58.710 (4) & 5/6/1 & 66.279 (4) \\
		\hline
	\end{tabular}
	\end{table*}


	\begin{table*}
		\centering
		\caption{CEC2011 results ($N_{\max}=50{,}000$, $100{,}000$): accuracy $\mathcal{E}$, rank $R$, combined score $S$, and ARRDE win/tie/loss (W/T/L) over 22 functions. Parentheses indicate ranks; bold denotes best.}		
	\label{2011}
		\TableFont
		\begin{tabular}{l|cccc|cccc}
			\hline
			Algorithm & \multicolumn{4}{|c|}{$N_{\max}=50000$} & \multicolumn{4}{|c}{$N_{\max}=100000$} \\
			& $\mathcal{E}$ & $R$ & $S$ & W/T/L & $\mathcal{E}$ & $R$ & $S$ & W/T/L \\
			\hline
			ARRDE & \textbf{0.100 (1)} & \textbf{2.276 (1)} & \textbf{100.000 (1)} & 0/22/0 & \textbf{0.066 (1)} & \textbf{2.697 (1)} & \textbf{100.000 (1)} & 0/22/0 \\
			LSHADE & 0.152 (2) & 3.478 (4) & 65.655 (3) & 14/5/3 & 0.094 (2) & 3.256 (4) & 76.553 (3) & 11/7/4 \\
			jSO & 0.164 (3) & 3.438 (3) & 63.467 (4) & 13/6/3 & 0.112 (4) & 3.230 (3) & 71.438 (4) & 9/11/2 \\
			LSHADE-cnEpSin & 0.168 (4) & 2.933 (2) & 68.439 (2) & 13/4/5 & 0.099 (3) & 2.701 (2) & 83.471 (2) & 6/8/8 \\
			j2020 & 0.415 (6) & 5.497 (6) & 32.722 (6) & 18/4/0 & 0.349 (6) & 5.495 (6) & 34.050 (6) & 18/3/1 \\
			NL-SHADE-RSP & 0.261 (5) & 4.398 (5) & 45.023 (5) & 17/4/1 & 0.186 (5) & 4.498 (5) & 47.828 (5) & 16/4/2 \\
			NL-SHADE-LBC & 0.587 (7) & 5.981 (7) & 27.530 (7) & 18/4/0 & 0.578 (7) & 6.123 (7) & 27.765 (7) & 18/4/0 \\
			\hline
		\end{tabular}
		\end{table*}

\begin{table*}[t]
\centering
\TableFont
		\setlength{\tabcolsep}{4pt}
		\caption{CEC2011 results ($N_{\max}=150{,}000$) with combined scores across all budgets: $\mathcal{E}$, $R$, $S$, and ARRDE win/tie/loss (W/T/L) over 22 functions. Parentheses indicate ranks; bold denotes best.}

	\label{2011-2}	
	\begin{tabular}{l|cccc|c}
			\hline
			Algorithm & \multicolumn{4}{|c|}{$N_{\max}=150000$} & \multicolumn{1}{|c}{Comb.} \\
			& $\mathcal{E}$ & $R$ & $S$ & W/T/L & $S_{\text{tot}}$  \\
			\hline
			ARRDE & \textbf{0.072 (1)} & 2.812 (2) & \textbf{98.614 (1)} & 0/22/0 & \textbf{100.000 (1)}  \\
			LSHADE & 0.085 (3) & 3.199 (3) & 84.913 (4) & 10/9/3 & 75.129 (3)   \\
			jSO & 0.083 (2) & 3.216 (4) & 85.993 (2) & 10/8/4 & 72.558 (4)  \\
			LSHADE-cnEpSin & 0.102 (4) & \textbf{2.734 (1)} & 85.089 (3) & 7/5/10 & 78.710 (2) \\
			j2020 & 0.317 (6) & 5.500 (6) & 36.192 (6) & 18/3/1 & 34.611 (6)  \\
			NL-SHADE-RSP & 0.151 (5) & 4.406 (5) & 54.809 (5) & 16/4/2 & 49.178 (5)  \\
			NL-SHADE-LBC & 0.568 (7) & 6.133 (7) & 28.612 (7) & 18/4/0 & 28.211 (7) \\
			\hline
	\end{tabular}
		\end{table*}

\begin{table*}[t]
\centering
\TableFont
		\setlength{\tabcolsep}{4pt}
		\caption{CEC2019 results ($N_{\max}=10^8$) with original CEC2019 rankings: $\mathcal{E}$, $R$, $S$, and ARRDE win/tie/loss (W/T/L) over 10 functions (7 at 10D; others at 9D, 16D, 18D).  $S_{2019}$ denotes the original CEC2019 score. Parentheses indicate ranks; bold denotes best.}
	\label{2019}
	\begin{tabular}{l|cccc|c}
			\hline
			Algorithm & \multicolumn{4}{|c|}{$N_{\max}=10^8$} & \multicolumn{1}{|c}{Orig.} \\
			& $\mathcal{E}$ & $R$ & $S$ & W/T/L & $S_{2019}$ \\
			\hline
			ARRDE & \textbf{0.009 (1)} & \textbf{3.086 (1)} & \textbf{100.000 (1)} & 0/10/0 &  83.000 (3)  \\
			LSHADE & 0.205 (7) & 4.856 (7) & 33.959 (7) & 5/4/1 &  63.280 (7)\\
			jSO & 0.110 (4) & 3.629 (3) & 46.594 (4) & 4/6/0 & 75.440 (4)  \\
			LSHADE-cnEpSin & 0.179 (6) & 4.731 (6) & 35.116 (6) & 5/5/0 & 64.560 (6)   \\
			j2020 & 0.009 (2) & 3.911 (4) & 87.700 (2) & 3/6/1 & \textbf{87.251 (1) } \\
			NL-SHADE-RSP & 0.127 (5) & 4.405 (5) & 38.557 (5) & 4/6/0 &  66.720  (5)  \\
			NL-SHADE-LBC & 0.079 (3) & 3.381 (2) & 51.273 (3) & 0/8/2 & 83.360 (2) \\
			\hline
	\end{tabular}
		\end{table*}

\subsubsection{CEC2022}

On CEC2022, which includes 10D and 20D problems under relatively large evaluation budgets that remain smaller than those of CEC2020, ARRDE again ranks first overall, as shown in Table~\ref{2022}. It achieves the highest combined score $S_{\text{tot}}$, and at 20D it attains the best results in all three metrics ($\mathcal{E}$, $R$, and $S$). At 10D, j2020 slightly surpasses ARRDE in the combined score $S$, but ARRDE still ranks first in $R$ and remains very close in accuracy $\mathcal{E}$. Across both dimensions, ARRDE also shows the strongest W/T/L profile, winning more pairwise comparisons than it loses over the 12 functions. It is also noteworthy that NL-SHADE-LBC, which ranked second in the official CEC2022 competition with a score very close to that of the winner, ranks fourth under the evaluation criteria adopted in this study. Taken together, these results indicate that ARRDE remains highly competitive in this low-dimensional setting even when compared with methods specifically designed for the recent CEC2022 benchmark regime.

\subsubsection{CEC2011}
The CEC2011 real-world suite represents a demanding regime in which many problems are high-dimensional, while the available evaluation budgets remain relatively small. Several problems have dimension 96, and some reach 120, 140, or 216, yet the budgets are limited to only $50{,}000$, $100{,}000$, and $150{,}000$ evaluations. For comparison, the CEC2017 benchmark prescribes $10{,}000 \times D$ evaluations, which would be much larger for problems of this scale. Tables~\ref{2011}--\ref{2011-2} therefore provide a useful test of whether the strong high-dimensional behavior observed on CEC2017 transfers to more structured real-world problems under tighter evaluation limits. Overall, the answer is largely positive: ARRDE ranks first under the combined score $S_{\text{tot}}$, while LSHADE-cnEpSin is the most competitive baseline across the three budgets.

At the budgets $N_{\max}=50{,}000$ and $N_{\max}=100{,}000$ (Table~\ref{2011}), ARRDE performs best in both accuracy and ranking, and therefore also attains the highest combined score. When the budget is increased to $150{,}000$, the comparison becomes closer. ARRDE still achieves the best accuracy-based score $\mathcal{E}$ and the best combined score $S^{(D)}$, whereas LSHADE-cnEpSin attains the best rank-based score $R$. This pattern suggests that LSHADE-cnEpSin performs better on a slightly larger number of individual problems, while ARRDE tends to obtain lower average errors on the problems where it performs well. Under the present robustness-oriented evaluation, that balance favors ARRDE.

The aggregated results over all three budgets reinforce this interpretation. ARRDE attains the highest overall combined score, followed by LSHADE-cnEpSin, while LSHADE and jSO form a weaker middle tier. j2020, NL-SHADE-RSP and NL-SHADE-LBC consistently occupy the last three positions, indicating that their strengths in low-dimensional, exploration-oriented settings do not transfer well to these predominantly high-dimensional, budget-constrained real-world problems.

A further point of interest is the relative behavior of jSO and LSHADE. On CEC2017, jSO typically ranks ahead of LSHADE, whereas on CEC2011 LSHADE attains lower average errors even when jSO remains slightly competitive in the rank-based score. This suggests that, for these real-world high-dimensional problems, LSHADE's more conservative search dynamics can be advantageous under tight evaluation budgets.

\subsubsection{CEC2019 100-Digit Challenge}

While CEC2011 represents very high dimensionality under restricted evaluation budgets, the CEC2019 100-Digit Challenge represents the opposite extreme. It consists mainly of low-dimensional problems, seven at 10D and the remainder at 9D, 16D, and 18D, but allows an extremely large evaluation budget. In the original competition, some methods used $N_{\max}>10^9$. In the present experiments, we adopt the still very large budget $N_{\max}=10^8$. This setting is therefore closer in spirit to CEC2020 than to CEC2017, since extensive search becomes more important than strict budget efficiency.

Under the robustness-oriented evaluation used in this study, ARRDE ranks first overall in Table~\ref{2019}. It achieves the lowest error $\mathcal{E}$, the best rank-based score $R$, and therefore the highest combined score $S=100$. The closest competitor is j2020, whose error is nearly identical to that of ARRDE but whose rank-based score is clearly weaker, which leaves it in second place overall. NL-SHADE-LBC ranks third: it is competitive in the rank-based measure and also performs well under the original CEC2019 score, but its much larger error prevents it from matching ARRDE or j2020 under the robustness-oriented aggregation.

The official CEC2019 score $S_{2019}$ leads to a different ordering. Under this metric, j2020 ranks first with 87.251, NL-SHADE-LBC ranks second with 83.360, and ARRDE ranks third with 83.000. This difference is consistent with the purpose of $S_{2019}$, which emphasizes high digit accuracy from the 25 best runs, whereas the evaluation adopted here is designed to reflect more stable overall behavior across runs.

\subsection{CEC2017, CEC2020, and CEC2022 under Varying \texorpdfstring{$N_{\max}$}{Nmax}}

While ARRDE performs strongly on all benchmark suites under the official evaluation budgets, practical optimization often requires robustness across a wide range of function-evaluation limits. Since algorithm rankings can change substantially when $N_{\max}$ deviates from competition settings~\citep{PIOTROWSKI2025101807}, we further assess the behavior of ARRDE and competing algorithms under varying budgets of the form $N_{\max}=500D, 1000D, \ldots, 10{,}000D$. This analysis is particularly relevant for real-world scenarios where evaluation budgets are frequently much smaller than those prescribed in CEC competitions.

\begin{figure*}[t]
	\centering
	
	\begin{subfigure}{0.32\textwidth}
		\centering
		\includegraphics[width=\linewidth]{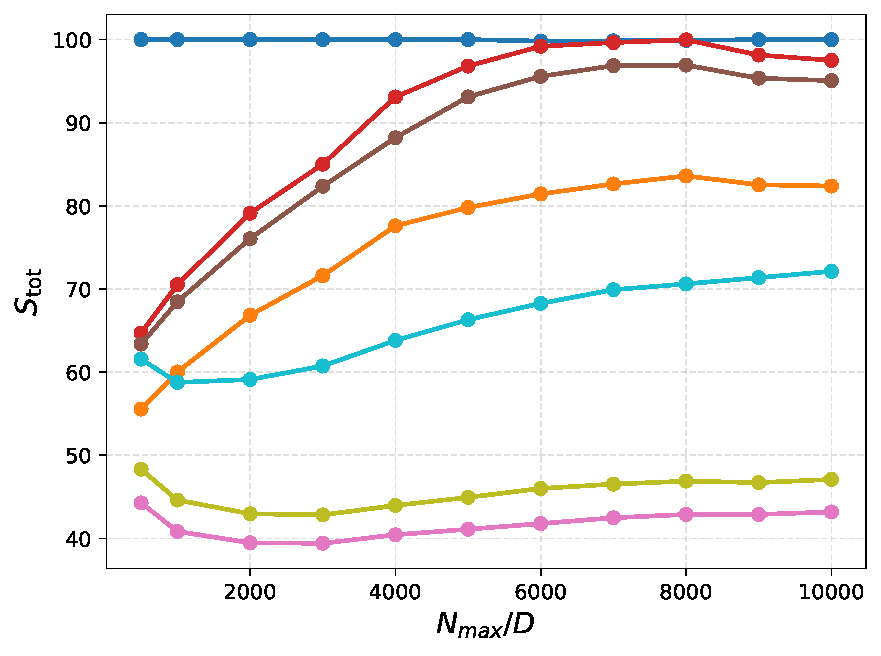}
		\caption{CEC2017}
		\label{fig:2017-s-vs-nmd}
	\end{subfigure}
	\hfill
	\begin{subfigure}{0.32\textwidth}
		\centering
		\includegraphics[width=\linewidth]{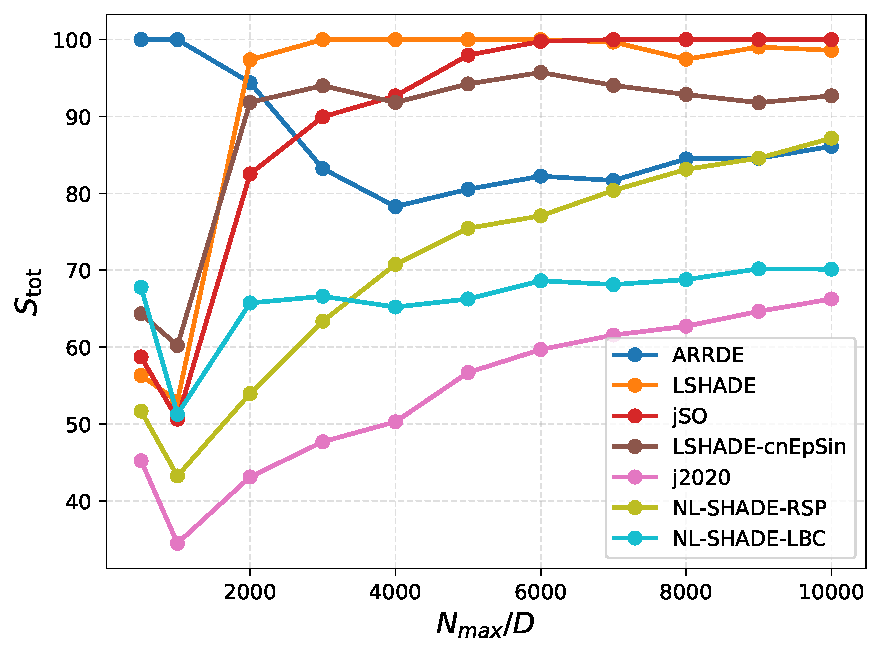}
		\caption{CEC2020}
		\label{fig:2020-s-vs-nmd}
	\end{subfigure}
	\hfill
	\begin{subfigure}{0.32\textwidth}
		\centering
		\includegraphics[width=\linewidth]{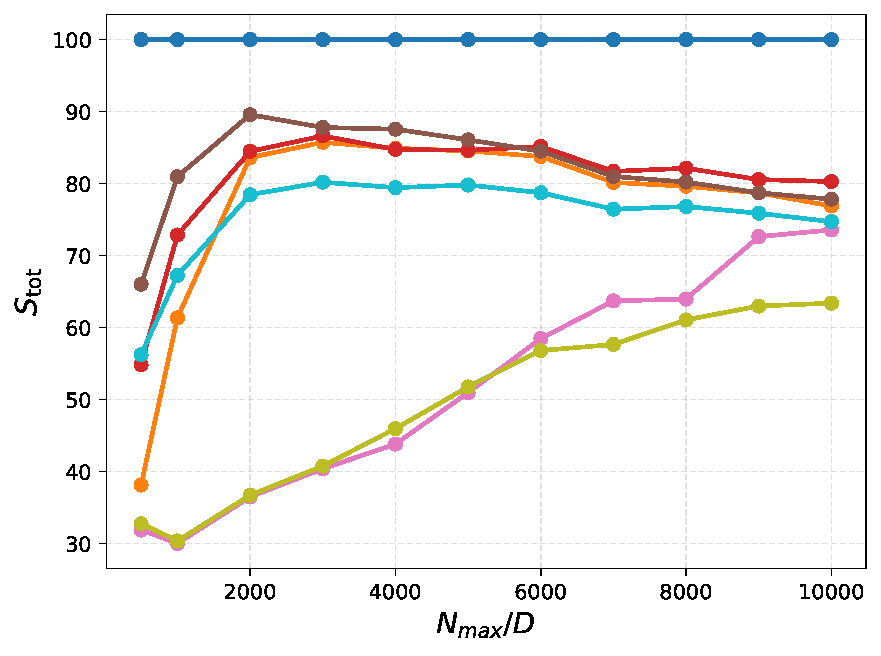}
		\caption{CEC2022}
		\label{fig:2022-s-vs-nmd}
	\end{subfigure}
	
	\caption{Combined score $S_{\text{tot}}$ as a function of normalized evaluation budget $N_{\max}/D$ for the CEC2017, CEC2020, and CEC2022 benchmark suites.}
	\label{fig:s-vs-nmd-all}
\end{figure*}

Figure~\ref{fig:s-vs-nmd-all} reports the combined score $S$ as a function of the normalized budget $N_{\max}/D$ for CEC2017, CEC2020, and CEC2022. On CEC2017 [Fig.~\ref{fig:s-vs-nmd-all}(a)], ARRDE maintains $S = 100$ across the entire budget range. The gap to jSO and LSHADE-cnEpSin decreases as the budget increases, but the relative ordering remains essentially unchanged. j2020 and NL-SHADE-RSP stay at the bottom throughout. Thus, on this suite the score differences vary with $N_{\max}/D$, whereas the ranking itself is stable.

A different pattern appears on CEC2020 [Fig.~\ref{fig:s-vs-nmd-all}(b)]. Here, the relative ordering changes substantially with $N_{\max}/D$. ARRDE performs best at small budgets, whereas j2020, NL-SHADE-LBC, and NL-SHADE-RSP improve steadily as the budget increases. LSHADE, jSO, and LSHADE-cnEpSin also benefit from additional budget and become more competitive over much of the intermediate range, roughly $3000 \leq N_{\max}/D \leq 10000$. Since ARRDE, NL-SHADE-RSP, and j2020 rank near the top under the official CEC2020 budget setting, as discussed in Section~\ref{sec:20}, this result illustrates that relative algorithm performance can change markedly across evaluation regimes.

On CEC2022 [Fig.~\ref{fig:s-vs-nmd-all}(c)], the behavior again resembles that of CEC2017. ARRDE consistently attains $S=100$ across all budget levels, and the relative ordering of the competing algorithms remains stable over the tested budget range. As in CEC2017, the score gaps vary with $N_{\max}/D$, but the ranking changes little.

Overall, the varying-budget results show that ARRDE remains strong across the tested budgets, while the stability of the relative ranking depends on the benchmark suite. The ordering is largely stable on CEC2017 and CEC2022, but changes substantially on CEC2020.

\section{Conclusion}
This paper presented the Adaptive Restart--Refine Differential Evolution (ARRDE) algorithm, a DE variant aimed at improving robustness across different problem characteristics and evaluation budgets. Built on jSO, ARRDE combines three design components: an adaptive restart--refine mechanism, including final refinement and local exclusion during restart; a nonlinear, dimension-dependent population-size schedule; and a budget-aware initialization rule for restricted-budget settings.

The experimental results show that restart--refine is particularly beneficial in low-dimensional, large-budget settings, while the proposed nonlinear reduction further improves cross-regime robustness without sacrificing the strong high-dimensional behavior of the jSO baseline. The budget-aware initialization rule provides an additional practical gain in restricted-budget settings and contributes to the overall robustness of the full ARRDE design. Across five benchmark suites, namely CEC2011, CEC2017, CEC2019, CEC2020, and CEC2022, as well as additional experiments over varying normalized budgets, ARRDE demonstrated strong and comparatively stable overall performance against a diverse set of baselines. The official suite-specific metrics and the proposed robustness-oriented metric do not always produce identical rankings, and some tuning was performed on CEC2017 and CEC2020; within those limitations, the results indicate that ARRDE is a competitive DE variant with a strong cross-suite robustness profile over the evaluated benchmark regimes.

A practical tradeoff of ARRDE is that, relative to the underlying baseline, it introduces additional design parameters. In the present study, these parameters were calibrated and tested over an extensive collection of CEC benchmark problems, including restricted-budget settings with small $N_{\max}$, so further retuning may often be unnecessary in similar regimes. Nevertheless, reducing the number of explicit design parameters while preserving the observed performance remains an important direction for future work. One promising route is to replace part of the current empirical parameterization with more fundamental design principles that can recover similar regime-adaptive behavior with a simpler parameter structure.

\bibliographystyle{unsrtnat}
\bibliography{ref}


\end{document}